# Automated Essay Scoring in Arabic: A Dataset and Analysis of a BERT-based System

Rayed Ghazawi[1] Edwin Simpson[2]

July 15, 2024

**Abstract** Automated Essay Scoring (AES) holds significant promise in the field of education, helping educators to mark larger volumes of essays and provide timely feedback. However, Arabic AES research has been limited by the lack of publicly available essay data. This study introduces AR-AES, an Arabic AES benchmark dataset comprising 2046 undergraduate essays, including gender information, scores, and transparent rubric-based evaluation guidelines, providing comprehensive insights into the scoring process. These essays come from four diverse courses, covering both traditional and online exams. Additionally, we pioneer the use of AraBERT for AES, exploring its performance on different question types. We find encouraging results, particularly for Environmental Chemistry and source-dependent essay questions. For the first time, we examine the scale of errors made by a BERT-based AES system, observing that 96.15% of the errors are within one point of the first human marker's prediction, on a scale of one to five, with 79.49% of predictions matching exactly. In contrast, additional human markers did not exceed 30% exact matches with the first marker, with 62.9% within one mark. These findings highlight the subjectivity inherent in essay grading, and underscore the potential for current AES technology to assist human markers to grade consistently across large classes.

## 1 Introduction

Essay writing is an important tool for developing and assessing students' cognitive abilities, including critical thinking, communication skills and depth of understanding [11,34]. However, as student numbers grow, marking essays by hand becomes impractical, discouraging the use of essay questions in education [6]. AES systems [32] aim to reduce the time needed to mark essays, by assessing both writing skills and cognitive outputs automatically, and can mitigate scoring biases and inconsistencies arising from teacher subjectivity [7]. Despite extensive research in English [38,25], AES for Arabic, the fourth most widely used Internet language [1], remains underexplored, with most efforts concentrated on scoring short, one or two-sentence answers [7]. With the abundant youth population in the Arab world, the education system faces challenges due to a shortage of teachers and the inability to provide individualized feedback to students [13]. In addition, the Arabic language differs from English in terms of grammar, structural rules, and the formulation of ideas, which prevents the application of scoring systems designed for English [13]. In this context, the development of an Arabic essay scoring system is an urgent necessity.

Previous research has predominantly leaned on feature engineering in conjunction with shallow models, yielding only moderate performance outcomes [5,17]. In contrast, the potential of pretrained models such as AraBERT [9], AraVec [35], and AraGPT-2 [10], which learn vector representations from extensive text corpora, remains largely untapped within the context of Arabic AES. These models have demonstrated notable efficacy in various domains, encompassing tasks like question-answering, named entity recognition, sentiment analysis, and even the automatic scoring of short answers [29,4]. A major barrier to further research is the lack of publicly available datasets: datasets used in prior studies are either inaccessible or consist only of one or two-sentence answers.

To address these gaps, this study introduces AR-AES dataset, which consists of Arabic essays each marked by two different university teaching professionals. This dataset was collected from undergraduate students across diverse disciplines, covering various topics and writing styles. We include ancillary information, such as the gender of the students (male and female students were taught separately), the specific evaluation criteria (rubrics) employed, and model answers for each question. The dataset comprises a total of 12 questions and 2046 essays, collected through both traditional and online examination methods, and encompasses substantial linguistic diversity, with a total length of 115,454 tokens and 12,440 unique tokens.

This study also pioneers the use of AraBERT in Arabic AES by conducting a series of experiments to assess AraBERT's performance on our dataset at different levels of granularity, from the complete dataset down to individual courses and questions. We also examined AraBERT's performance based on gender, exam type (traditional or online), and essay type (argumentative, narrative, source-dependent). AraBERT excelled when trained on several questions from the same course, achieving a Quadratic Weighted Kappa (QWK) score of 0.971 in Environmental Chemistry. However, its performance was comparatively lower when trained specifically for certain types of question, with the lowest QWK observed for narrative questions.

Our analysis goes beyond previous work on AES, by assessing the proximity of the model's predictions to the grades assigned by the first marker, to gauge the scale of its errors. The predictions matched exactly for 79.49% of answers, with 95% of predictions having no more than one mark difference to the first human mark (out of a total of five marks). In contrast, the question with highest agreement between the first and second human markers had only 30.3% exact agreement, with differences greater than one mark for 37.1% of the answers. This suggests that AraBERT-based AES is sufficiently capable to assist human markers and could help detect inconsistencies between individuals in a marking team.

[1] Internet World State ranking, March 2020, www.internetworldstats.com

In summary, our study presents a comprehensive approach to Arabic AES, introducing an open-source dataset with clear annotation guidelines and quality control, leveraging AraBERT, and providing a novel investigation of the scale of AraBERT AES errors. We commit to making our code, data, and marking guidelines publicly accessible upon acceptance.

## 2 Related Works

Several AES datasets have been released in Chinese [18], Indonesian [1], and English, including the ASAP dataset[2] that has catalysed English AES research [33,37], including a new state-of-the-art BERT-based approach [38]. However, there is no previous publicly available dataset containing Arabic essays and marks, as existing work is limited to short answers [3]. Our study addresses this gap by presenting a comprehensive dataset for Arabic AES.

Arabic AES research encompasses approaches such as linear regression [5], Latent Semantic Analysis [2], Support Vector Machines [7], rule-based systems [6], naïve Bayes [3], and optimization algorithms like eJaya-NN [17]. However, these studies predominantly rely on feature engineering, using surface features that are unable to comprehensively capture the semantic nuances and structural intricacies inherent in essays. These approaches provide only limited consideration for word order, primarily revolving around word-level or grammatical features. More recent pretrained transformer models, such as BERT [15], alleviate these issues but have not previously been harnessed for Arabic AES. Here, we develop the first AES system using AraBERT to analyse the effect of different question types on a modern text classifier. We also go beyond previous analyses of model performance by evaluating the magnitude of errors in the models' predictions, as large errors could have a greater impact on students.

## 3 Arabic language challenges

NLP systems face several distinct challenges when processing Arabic, which motivate the development of bespoke tools and language resources, including benchmark datasets.

**Linguistic Complexity:** Arabic exhibits complex sentence structures with many syntactic and stylistic variations, an extensive vocabulary, and the frequent use of rhetorical devices [8]. Arabic, for instance, has many ways to express the concept of "going" depending on who is doing the action, when, and whether the action is done in a habitual or momentary sense. For example, يذهب (he goes), سأذهب (I will go), كان يذهب (he used to go), and يذهبان (they (two) go).

[2] www.kaggle.com/c/asap-aes

This complexity can make it hard for an AES system to recognise variations of the same concept.

**Complex Morphology:** Arabic features intricate morphology, encompassing a wide range of inflection and derivational systems [19]. Words in Arabic can have multiple forms based on factors such as tense, gender, number, and case, and the form of a single letter also varies. For instance, the letter س ('S'), looks like (سـ) at the beginning of a word (سحاب“Cloud”), like (ـسـ) in the middle as in (مستشفى, “Hospital”), and like (ـس) at the end as in (شمس, “Sun”). This complexity adds to the difficulty of stemming, tokenization, and lemmatization operations [24]. As another example, the Arabic root word for “write” is كتب, from which we can derive various words like كاتب (“writer”), مكتوب (“written”), كتاب (book), كتبت (“I wrote”), يكتب (“he writes”), etc. The challenge for AES systems here lies in recognizing these words as related.

**Non-Standard Orthography:** Arabic text follows complex rules for letter representation, including ligatures and diacritics that influence pronunciation, word comprehension, and meaning [21,36]. NLP systems face challenges in handling these orthographic differences and the absence of diacritics in unvocalised text. For example, the word محبوبة (“loved or popular”) could be written as محبوبه in casual writing without the ending diacritic.

**Lack of Resources:** Arabic suffers from limited linguistic resources, such as preprocessing tools for dealing with the language complexities described above, and a lack of available datasets [27,23], which hampers the development of NLP models. A particular need is for bespoke tools to deal with the right-to-left text direction, which creates additional complexities for mixed-language content [12,24]. This study contributes a labelled dataset in Arabic, which will enable further development of Arabic NLP systems.

**Ambiguity and Polysemy:** Arabic words often possess multiple meanings and interpretations, making it challenging to disambiguate them [16]. For example, the word جمل in Arabic can mean “camel” or “sentence” depending on context. Contextual analysis becomes crucial for accurately determining the intended meaning [23,31]. This aspect presents a challenge in various NLP tasks, including named entity recognition, sentiment analysis, and machine translation.

Despite these challenges, substantial advancements have been made in Arabic NLP in recent years, including language models and tools specifically designed for Arabic. This study hopes to contribute to this effort.

## 4 The AR-AES Dataset

The AR-AES[3] dataset is intended for both training and evaluating Arabic AES systems, and covers essays written by both male and female undergraduate students from three different university faculties, with a range of different question types, a mix of traditional face-to-face and online exams, and marks from multiple human markers. As part of the dataset, we include clear and detailed marking criteria along with model answers for each question. This diversity will enable researchers to explore the suitability of AES systems for different types of essays, exam types, or student cohorts.

[3] The dataset and appendices can be accessed via the following link: `https://osf.io/dp2nh/?view_only=4ac6373c60214ea6952855f81507fec7`

*Data Collection:* To compile a diverse dataset, we first selected multiple undergraduate courses across various departments at Umm Al-Qura University, as shown in Table 1. Students' writing skills vary depending on their academic disciplines [39], due to differing objectives, domain-specific terminology, and research formulation methodologies. Additionally, factors like gender and academic level contribute to differences in writing [22,26], particularly considering that the genders are taught separately. Therefore, to facilitate testing of AES systems across various subjects and writing styles, we collected essay responses from diverse academic levels, and male and female genders.

Ensuring diversity in question types within the dataset was vital for evaluating the model's performance across various essay categories.

To bolster dataset diversity, we employed both traditional (in-person) and online exams through distance learning. Traditional exams occurred on specific dates within campus halls or laboratories, subjecting students to controlled conditions that minimized opportunities for academic misconduct. Conversely, online exams required students to submit essay responses exclusively via content management platforms. These exams shared time limits with traditional exams but did not mandate physical presence on campus. Online exams can reduce stress levels [20], granting students greater freedom in providing answers and potentially allowing access to course content during the exam. For both kinds of exam, answers were typed and submitted electronically, eliminating the need to convert handwritten answers to digital format. These essays were part of the students' compulsory assessment within the midterm exams for their respective courses, and they volunteered to provide their essays for our dataset.

| Course | Faculty | Semester | Exam Type | No. Groups | Gender | No. Students | Question ID | Essay Type | Answer Length Max | Answer Length Min | Score Range Min | Score Range Max |
|---|---|---|---|---|---|---|---|---|---|---|---|---|
| Introduction to Info Science | Computing | 1 | Traditional | 3 | Male | 151 | | All questions | 298 | 2 | 0 | 5 |
| | | | | | | | Q1 | Narrative | 298 | 7 | 0 | 5 |
| | | | | 2 | Female | 128 | Q2 | Argumentative | 164 | 2 | 0 | 5 |
| | | | | | | | Q3 | Source Dependent | 61 | 4 | 0 | 5 |
| Management Info Systems | Business Administration | 5 | Traditional | 2 | Male | 181 | | All questions | 512 | 16 | 0 | 10 |
| | | | | | | | Q4 | Narrative | 512 | 29 | 0 | 10 |
| | | | | | | | Q5 | Narrative | 212 | 29 | 0 | 10 |
| | | | | | | | Q6 | Source Dependent | 171 | 16 | 0 | 5 |
| Environmental Chemistry | Applied Science | 7 | Online | 2 | Male | 116 | | All questions | 422 | 8 | 0 | 5 |
| | | | | | | | Q7 | Narrative | 422 | 25 | 0 | 5 |
| | | | | | | | Q8 | Argumentative | 116 | 9 | 0 | 5 |
| | | | | | | | Q9 | Source Dependent | 92 | 8 | 0 | 5 |
| Biotechnology | Applied Science | 6 | Online | 2 | Male | 106 | | All questions | 575 | 11 | 0 | 5 |
| | | | | | | | Q10 | Source Dependent | 357 | 13 | 0 | 5 |
| | | | | | | | Q11 | Argumentative | 538 | 11 | 0 | 5 |
| | | | | | | | Q12 | Source Dependent | 575 | 13 | 1 | 5 |

Table 1: Course summary, including the semester in which the exam was taken (out of 8 semesters in an undergraduate degree), the number of groups taught at separate times (No. Groups), and answer lengths (number of tokens).

| Rubric-based evaluations | Score |
|---|---|
| وأشكالها ودورها بالبيانات التعريف على الطالب قدرة<br>The student's ability to define data, its role and forms | 1 |
| استخدامها وأوجه ونشأتها بالمعلومات التعريف على الطالب قدرة<br>The student's ability to identify information, its origins, and its uses | 1 |
| والمعلومات البيانات بين الفرق استنتاج على الطالب قدرة<br>The student's ability to deduce the difference between data and information | 2 |
| صلة ذات واقعية بأمثلة والمعلومات للبيانات شرحه تعزيز على الطالب قدرة<br>The student's ability to reinforce their explanation of data and information with relevant examples | 1 |
| **Final Score** | 5 |

Table 2: Example marking criteria set by the course director for Q1.

*The Annotation Task:* Course directors equipped markers with detailed guidelines for scoring individual criteria and determining the final score. Table 2 shows an example of the criteria employed for assessing Question 1, which prompts students to "Explain in detail the difference between the terms 'data' and 'information', supplementing their answers with examples of each type". For an exhaustive overview of the Scoring Criteria, see Table A.3. This structured approach facilitates the identification of essay strengths and weaknesses. Nonetheless, essay scoring is a subjective process, and different markers could still assign varying scores to the same essay. Relying solely on a single score, as in the ASAP datasets, could therefore limit the accuracy and comprehensiveness of model testing. Hence, to investigate the degree of variation between human markers, and provide a point of comparison for the agreement level between human markers and AES systems, we collected two scores for each essay. For all questions, the first marker is the course provider; second markers were members of the same faculty who were familiar with the course content. In total, a team of 9 faculty members formulated, prepared, and scored the exams.

*Quality control:* Firstly, to guarantee the quality of essay questions, individual meetings were conducted with faculty members responsible for each course. The course directors were provided with criteria for formulating essay questions, and then the proposed questions were verified by the authors of this paper against these criteria, and revised if they did not meet the criteria. The following criteria were presented as to faculty members for formulating the essay questions:

1. **Clear Objectives:** Each question should have a clear objective aimed at assessing a specific cognitive skill, such as analysis, synthesis, or evaluation. This clarity helps students focus on comprehending the question and providing the required answer directly.
2. **Relevance:** Ensure that the question directly relates to the course content and learning objectives.
3. **Explicit Terminology:** During the question formulation process, it is advisable to incorporate explicit terminology relevant to the course content.
4. **Clarity and Simplicity:** Emphasis should be placed on crafting questions that are straightforward, unambiguous, and include a comprehensive outline of the expectations for the answer. This approach encourages students to provide concise and easily evaluated responses.

5. **Linguistic and Grammatical Accuracy:** When composing questions, meticulous attention should be given to linguistic and grammatical aspects. Ensuring that questions are free of grammatical errors prevents unintended alterations in question meaning.
6. **Alignment with Learning Outcomes:** Align each question with the specific learning outcomes you want to assess.
7. **Fairness:** Craft questions that offer all students an equal opportunity to demonstrate their knowledge and skills.
8. **Grading Guide:** For each question, a guide should be developed to communicate the correct answer structure and the specific criteria for achieving higher grades, clarifying the grading process.

In addition, special instructions were developed for online exams to prevent cheating. These measures included restricting exam access to one hour on the Blackboard platform and requiring students to have their cameras on throughout the exam. Students were explicitly instructed not to engage in chat conversations or pose questions during the examination. Any inquiries or concerns related to the test were to be addressed only after the examination had concluded.

*Dataset Statistics:* In total, we collected and labelled 2046 essays, with statistics shown in Table 3. Table 1 shows notable variations in answer lengths, measured in tokens, across different question types, and between online and traditional exam types, with online exam responses generally being longer across most questions. The class distribution is illustrated in Figure B.1.

| **Course Name** | **Questions Count** | **Essay Count** | **Gender** | | **Exam type** | |
|---|---|---|---|---|---|---|
| | | | **M** | **F** | **Traditional** | **Online** |
| Introduction to Information Science | 3 | 837 | 453 | 384 | 837 | |
| Management information systems | 3 | 543 | 543 | | 543 | |
| Environmental chemistry | 3 | 348 | 348 | | | 348 |
| Biotechnology | 3 | 318 | 318 | | | 318 |
| **Total** | **12** | **2046** | **1662** | **384** | **1380** | **666** |

Table 3: The number of collected essay responses for each course.

## 5 Experimental Setup

The AraBERT model has consistently demonstrated state-of-the-art performance in various Arabic NLP tasks, including the automatic scoring of short answers [29], but its application to AES remains unexplored. Thus, this study's primary goal is to assess AraBERT's performance in AES and its ability to handle longer Arabic texts. Additionally, we aim to investigate whether performance varies depending on factors such as the subject, question type, exam type, or gender.

*Data Preprocessing:* We removed punctuation, hashtags, URLs, excess letter repetitions, emoticons, superfluous spaces, numbers, and diacritics, and normalized specific Arabic characters to their standard forms (e.g., ة > ه - ى > ي - أ إ اُ ا > ا - ئ ؤ > ء). We applied the ISRI Stemmer, in the manner of previous work [29], to simplify Arabic text by reducing words to their roots to minimise vocabulary diversity. We employed the AraBERT tokenizer, and sequences exceeding 512 tokens were truncated. Most essays fit this limit, except four from the Biotechnology course, exceeding up to 575 words.

*Model Design:* AraBERT is a variant of BERT that was pretrained on a substantial Arabic text dataset [9] and can be fine-tuned for specific tasks with minimal additional training data, reducing the time and resources needed for model development and deployment. This study used the large AraBERT configuration, featuring 12 encoder blocks, 1024 hidden dimensions, 16 attention heads, 512 sequence length, and 370 million parameters. To leverage AraBERT's pre-trained capabilities, we added a standard classification head on top of it, consisting of a single fully-connected layer. This approach follows the *HuggingFace Transformers* library conventions, providing a simple yet effective method for fine-tuning. This design choice allows us to efficiently map AraBERT's high-dimensional representations to target classification labels, ensuring both computational feasibility and high accuracy. Notably, this study marks the first application of AraBERT in automatic Arabic essay-scoring tasks.

*Model Training:* The system aims to assist the course presenter (first annotator), so the model was trained only on the labels provided by that person. To ensure comparability across questions, we normalized all scores in the dataset to the range 0 to 5. Specifically, for questions with scores originally ranging from 0 to 10 (Q4 and Q5), we divided the scores by 2 to align them with the score range used for other essays. We trained the model once on the complete dataset, as well as separately for each course and each question. We also trained the model separately on male and female essay responses for the Introduction to Information Science course (each gender was taught separately by different instructors), and on traditional and online essay responses, to observe differences in model performance that could affect each group differently.

For each of these experiments, we divided the answers randomly into training, validation and test sets (split 70/15/15). We trained using Adam optimiser and the hyperparameters, including batch size, dropout rate (0.2), and the number of training epochs, were tuned on the validation set for each experiment, as detailed in Table A.2[4]. Given the dataset's imbalanced nature, we employed class weights to give equal weight to each class in the dataset by assigning proportionally higher weights to instances from smaller classes. The distribution of classes for each question is illustrated in Figure B.1[5].

[4] Appendices: https://osf.io/dp2nh/?view_only=4ac6373c60214ea6952855f81507fec7

*Evaluation Metrics:* We adopted Quadratic Weighted Kappa (QWK) and F1 score as evaluation metrics. QWK, an extension of Cohen's $\kappa$, gauges the level of agreement between the scoring outcomes of two assessors [14]. This metric is commonly employed in AES evaluation because, unlike accuracy and F1 score, $\kappa$ considers chance agreement, providing a more reliable measure of rating concordance [28]. Moreover, QWK accommodates the ordinal nature of classes, crucial to essay scoring, and employs quadratic weights to reflect class rank order, a nuance unaddressed by accuracy and F1 scores. QWK is computed by:

$$QWK = 1 - \frac{\sum_{i,j} w_{i,j} O_{i,j}}{\sum_{i,j} w_{i,j} n_{i,1} n_{j,2}}, \tag{1}$$

where $w_{i,j} = \frac{(i-j)^2}{(N-1)^2}$ is the weight between mark $i$ and mark $j$, $N$ is the number of marks available, $O_{i,j}$ is the number of observations where the first assessor gave mark $i$ and the second assessor gave mark $j$, and $n_{i,k}$ is the number of times that assessor $k$ gave mark $i$.

## 6 Results

We first evaluated the AraBERT model on the entire dataset to gauge its performance when trained with more data and a variety of questions. Then, we trained and evaluated models using data from each course, individual question, question type, student gender, and exam type, to identify the kind of scenarios where the AES system could be more effective.

The results are shown in Table 4. On the complete dataset, the model achieved a QWK score of 0.884 and an F1 score of 0.78, but this was not the highest score, suggesting that while the larger training set may benefit this combined model, some essay types are more amenable to AES than others. For instance, the model performance in the Environmental Chemistry course exceeded that of the entire dataset, even though this course included responses in Arabic mixed with English terms. Among the different courses, performance was weakest on Management Information Systems, potentially due to the complexity of the material or student responses. The Management Information Systems course had approximately 4469 unique words (in extended answers), while Environmental Chemistry had around 2702 unique words (in restricted answers). This difference suggests that the Management Information Systems course featured more open-ended essays compared to the Environmental Chemistry course, where answers were more source-dependent and controlled, making them easier for the model to evaluate. Within Management Information Systems, two questions were narrative, which tend to be open-ended, possibly contributing to the model's lower performance in this course.

Compared to Biotechnology, performance on the Information Science course was weaker, despite its larger training set. We investigated whether this discrepancy may be attributed to the students' use of informal language, considering that this course is a first-semester offering for first-year undergraduates, while the Biotechnology course is taken in the second semester of the third year. We computed the perplexity [30] of students' answers for each course, finding that Introduction to Information Science had a high perplexity score of 14.87 compared to Management Information System (1.77), Environmental Chemistry (1.5), and Biotechnology (1.68). This suggests that the AraBERT model was less suitable for modelling the Introduction to Information Science answers, and that the language differs from that used in other courses.

| The Experiment | Unique words | F1 | QWK |
|---|---|---|---|
| The Entire Dataset | 12440 | 0.78 | 0.884 |
| Introduction to Information Science | 3953 | 0.61 | 0.788 |
| Management Information System | 4469 | 0.59 | 0.779 |
| Environmental chemistry | 2702 | 0.95 | 0.971 |
| Biotechnology | 4241 | 0.85 | 0.953 |
| Question 1 | 1922 | 0.59 | 0.887 |
| Question 2 | 1906 | 0.47 | 0.733 |
| Question 3 | 938 | 0.82 | 0.870 |
| Question 4 | 2331 | 0.82 | 0.833 |
| Question 5 | 1878 | 0.85 | 0.841 |
| Question 6 | 978 | 0.95 | 0.942 |
| Question 7 | 1801 | 0.33 | 0.425 |
| Question 8 | 767 | 0.88 | 0.791 |
| Question 9 | 507 | 0.91 | 0.979 |
| Question 10 | 772 | 0.77 | 0.902 |
| Question 11 | 1787 | 0.57 | 0.843 |
| Question 12 | 2483 | 0.76 | 0.838 |
| Female | 2723 | 0.59 | 0.741 |
| Male | 3033 | 0.53 | 0.715 |
| Traditional Exam | 7506 | 0.57 | 0.758 |
| Online Exam | 6355 | 0.72 | 0.929 |
| Narrative (Q1,Q4,Q5,Q7) | 6790 | 0.45 | 0.693 |
| Argumentative (Q2,Q8,Q11) | 3863 | 0.64 | 0.732 |
| Source Dependent (Q3,Q6,Q9,Q10,Q12) | 4667 | 0.73 | 0.889 |

Table 4: Comparison of AraBERT models trained on different question subsets.

Overall, the model performed best with source-dependent questions, where language is more constrained, and worst with narrative questions, which were the most open-ended, reflected in the higher number of unique words shown in Table 4. The model also performed better with online, rather than traditional in-person exams. Splitting the Introduction to Information Science questions by gender resulted in superior performance when predicting female students' marks, which may reflect different teaching or learning styles, as male and female students are taught separately by different lecturers.

*Magnitude of Errors:* It is important to consider the scale of errors that the model makes: if the system predicts marks that are much lower or higher than the human marker, students could be unfairly penalised or rewarded for poor-quality work. We therefore assess the deviations between predictions and correct scores using a confusion matrix, as shown in Figure 1. The pattern is similar across courses. The majority of errors involved overestimations, with 10% of cases resulting in a one-mark overestimation. Underestimations were less frequent, occurring in 6% of cases with a one-degree reduction. Exact matches were 12% higher for Environmental Chemistry than Introduction to Information Science. Examining the confusion matrix for each essay type (Figure 2), one-mark overestimates occur noticeably more in narrative essays, while source-dependent essay predictions match the human marker's grade in 87% of cases.

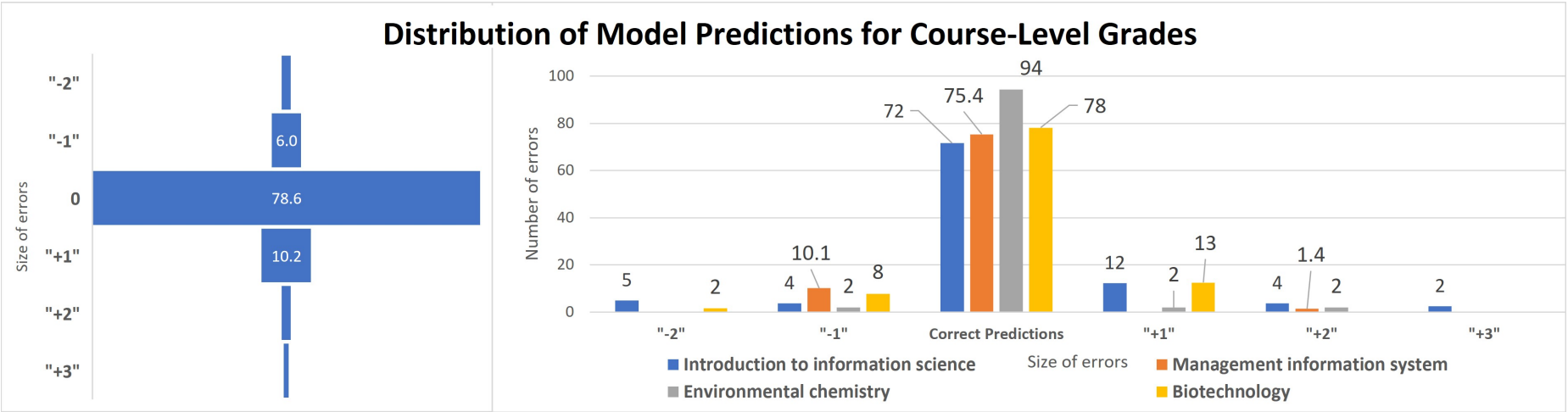

Figure 1: Distribution of model predictions for course-level grades.

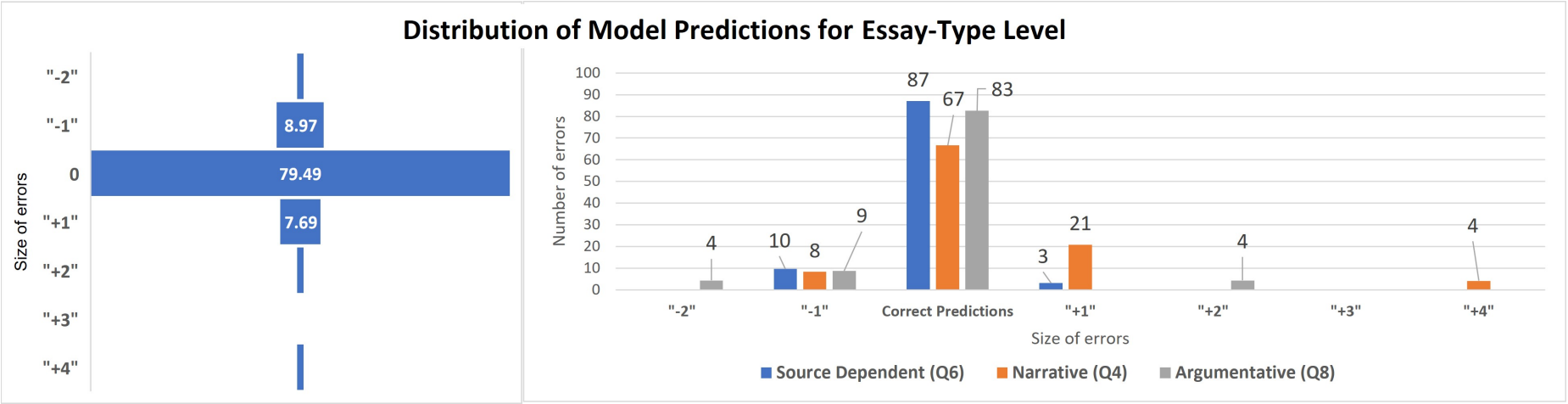

Figure 2: Distribution of Model Predictions for Essay-type Level.

## 7 Human Agreement

Consistent grading in education and assessment is crucial to maintain fairness and objectivity. Here, we assess the consistency of grades assigned by different human assessors across each question type and compare the level of agreement between human markers to the model's performance. We examine agreement for two courses: Introduction to Information Science and Information Systems Management, for a total of six questions (Q1 to Q6) and show results in Table 5.

The highest agreement among human assessors was observed in Q3 (source-dependent), where they provided the same grade in 30.3% of cases out of 279 responses. Negative differences were far more frequent than positive, meaning that second markers tended to mark more harshly than the course directors. Conversely, the lowest agreement among human assessors was found in Question 4 (narrative), which has notably more cases of disagreement by 3 or more marks.

When compared with the performance of our models, which were trained with the gold standard marks of the original markers, we see that the disparity in second marker's assessments often exceeds the error rate of the automated system. This suggests that the model may be an effective way to assist a human marker or could help to ensure consistency between multiple markers.

| | Correlation | QWK | Question Type | -5 | -4 | -3 | -2 | -1 | Matching | 1 | 2 | 3 | 4 | 5 |
|---|---|---|---|---|---|---|---|---|---|---|---|---|---|---|
| Question 1 | 0.574 | 0.543 | Narrative | 0 | 1.5 | 4.8 | 20.4 | 25.2 | 23.3 | 13.3 | 8.1 | 3.3 | 0 | 0 |
| Question 2 | 0.639 | 0.618 | Argumentative | 0 | 1.9 | 6.7 | 16.7 | 22.6 | 25.6 | 17.8 | 7.0 | 1.1 | 0.7 | 0 |
| Question 3 | 0.775 | 0.690 | Source dependent | 0 | 0.4 | 9.0 | 25.1 | 25.1 | 30.3 | 7.5 | 1.9 | 0.7 | 0 | 0 |
| Question 4 | 0.577 | 0.174 | Narrative | 0.6 | 2.6 | 12.3 | 19.5 | 24.7 | 20.8 | 9.1 | 7.1 | 1.3 | 1.9 | 0 |
| Question 5 | 0.834 | 0.252 | Narrative | 1.3 | 0 | 3.9 | 16.2 | 24.0 | 31.8 | 21.4 | 7.1 | 0 | 0.6 | 0 |
| Question 6 | 0.734 | 0.665 | Source dependent | 0 | 0 | 0 | 3.3 | 12.5 | 27.6 | 50.7 | 5.3 | 0.7 | 0 | 0 |

Table 5: The extent of agreement and discrepancy between the scores of the two human assessors is compared, in addition to the correlation, and QWK.

## 8 Conclusions and Future work

In this paper, we introduced AR-AES, the first publicly-available Arabic AES dataset, consisting of 2046 undergraduate essays with model answers, marking criteria, and scores from multiple markers. We also developed and evaluated an AES system using AraBERT, and demonstrated promising performance, particularly on source-dependent essays in domains such as Environmental Chemistry. Our analysis showed that agreement between our model and gold standard marks is higher than agreement among human markers, suggesting a role for AES in ensuring consistency as well as increasing marking efficiency.

There are numerous avenues for future work, such as exploring the adaptation of state-of-the-art techniques from the English AES field to the domain of Arabic AES, such as the multi-scale approach of [38]. In addition to model exploration, future research should also focus on integrating AES systems into the essay grading process effectively, and addressing students' and teachers' concerns about automated systems. This includes designing a process for identifying and rectifying errors, and ensuring that human teachers retain control while being assisted in grading a large set of essays. This area holds significant potential for enhancing the efficiency and accuracy of essay scoring, particularly in universities with limited teaching resources. We also see value in expanding our dataset with essays from a wider range of courses and educational institutions, thereby enhancing the robustness and versatility of our model, and investigating other aspects of student diversity beyond subject and gender. Our approach may also provide a template for AES data collection in other languages.

## Acknowledgements

We would like to thank all students and lecturers at Umm Al-Qura University who have kindly participated in this study. Experiments in this paper were carried out using the computational facilities of the Advanced Com- puting Research Centre, University of Bristol (http://www.bristol.ac.uk/acrc/).

# Appendix A. Tables

| Course Name | Question ID | The Questions | The Questions in Arabic |
|---|---|---|---|
| Introduction to Information Science | 1 | Explain in detail the difference between the terms data and information and reinforce your answers with examples for each type? | اشرح بشكل مفصل الفرق بين كلا من مصطلحي البيانات والمعلومات مع تعزيز إجاباتك بأمثلة لكل نوع؟ |
| | 2 | Explain in detail the role of the increase in subspecialties and the increase in topics influencing the information revolution (explosion)? | أشرح باستفاضة دور زيادة التخصصات الدقيقة وتزايد الموضوعات في التأثير على ثورة انفجار المعلومات؟ |
| | 3 | Through what you learned in the course, mention the comprehensive definition of the term information science? | من خلال ما تعلمته ضمن المقرر الدراسي أذكري التعريف الشامل لمصطلح علم المعلومات؟ |
| Management information systems | 4 | The administrative levels' tasks, roles, and duties differ in management, so explain in detail the difference between the roles and tasks of the different administrative levels while strengthening your answer with examples? | تختلف مهام وأدوار وواجبات المستويات الإدارية في الإدارة لذلك أشرح بشكل مفصل الاختلاف بين أدوار ومهام المستويات الإدارية المختلفة مع تعزيز اجابتك بأمثلة؟ |
| | 5 | Mention three main benefits of cloud computing from a business management perspective with an explanation? | أذكر ثلاثة من الفوائد الرئيسية للحوسبة السحابية من منظور إدارة الأعمال مع الشرح؟ |
| | 6 | Through what you have learned in the course, mention the comprehensive definition of the term information technology and reinforce your answer with examples? | من خلال ما تعلمته ضمن المقرر الدراسي أذكري التعريف الشامل لمصطلح علم المعلومات؟ |
| Environmental chemistry | 7 | Talk about the layers of the atmosphere, mentioning the height and temperature of each layer? | تحدث عن طبقات الغلاف الجوي مع ذكر ارتفاع كل طبقة ودرجة الحرارة فيها؟ |
| | 8 | What do you think about the importance of the ozone layer? | ما رأيك في أهمية طبقة الأوزون؟ |
| | 9 | What is the scientific definition of environmental chemistry? | ما هو التعريف العلمي لكيمياء البيئة؟ |
| Biotechnology | 10 | Define the term biotechnology? | عرف مصطلح التقنية الحيوية؟ |
| | 11 | Discuss whether eating genetically modified fruits is healthy or not? | ناقش تناول الفواكه المعدلة الوراثية صحي أم لا؟ |
| | 12 | Mention five of the applications of biotechnology in the medical field with explanation? | عدد خمسة من تطبيقات التقنية الحيوية في المجال الطبي مع الشرح؟ |

Table A.1: List of Questions Used in Each Course to Collect Essay Answers.

| Experiment Type | The Dataset | Batch Size | Gradient Steps | N. Essays | Training Size | Val Size | Test Size | Early Stop[5] |
|---|---|---|---|---|---|---|---|---|
| The Entire Dataset | - | 8 | 4 | 2046 | 1432.2 | 306.9 | 306.9 | 24 |
| Courses | Introduction to Information Science | 22 | 4 | 837 | 585.9 | 125.55 | 125.55 | 49 |
| | Management Information System | 9 | 4 | 543 | 380.1 | 81.45 | 81.45 | 15 |
| | Environmental Chemistry | 12 | 4 | 348 | 243.6 | 52.2 | 52.2 | 33 |
| | Biotechnology | 8 | 4 | 318 | 222.6 | 47.7 | 47.7 | 23 |
| Questions | Q1 | 12 | 6 | 279 | 195.3 | 41.85 | 41.85 | 10 |
| | Q2 | 12 | 4 | 279 | 195.3 | 41.85 | 41.85 | 43 |
| | Q3 | 128 | 2 | 279 | 195.3 | 41.85 | 41.85 | 13 |
| | Q4 | 8 | 4 | 181 | 126.7 | 27.15 | 27.15 | 47 |
| | Q5 | 28 | 4 | 181 | 126.7 | 27.15 | 27.15 | 44 |
| | Q6 | 42 | 4 | 181 | 126.7 | 27.15 | 27.15 | 10 |
| | Q7 | 12 | 4 | 116 | 81.2 | 17.4 | 17.4 | 10 |
| | Q8 | 8 | 2 | 116 | 81.2 | 17.4 | 17.4 | 27 |
| | Q9 | 8 | 2 | 116 | 81.2 | 17.4 | 17.4 | 44 |
| | Q10 | 16 | 2 | 106 | 74.2 | 15.9 | 15.9 | 60 |
| | Q11 | 8 | 8 | 106 | 74.2 | 15.9 | 15.9 | 42 |
| | Q12 | 8 | 8 | 106 | 74.2 | 15.9 | 15.9 | 47 |
| Gender | Female | 16 | 2 | 384 | 268.8 | 57.6 | 57.6 | 25 |
| | Male | 64 | 2 | 453 | 317.1 | 67.95 | 67.95 | 10 |
| Exam Type | Traditional Exam | 8 | 4 | 1380 | 966 | 207 | 207 | 22 |
| | Online Exam | 8 | 8 | 666 | 466.2 | 99.9 | 99.9 | 14 |
| Question Type | Narrative (Q1, Q4, Q5) | 16 | 2 | 757 | 529.9 | 113.55 | 113.55 | 18 |
| | Argumentative (Q2, Q8, Q11) | 16 | 2 | 501 | 350.7 | 75.15 | 75.15 | 24 |
| | Source Dependent (Q3, Q6, Q9, Q10, Q12) | 8 | 8 | 788 | 551.6 | 118.2 | 118.2 | 18 |

Table A.2: Summary of Experimental Setup and Hyperparameters.

| N | The Question | Potential Mark | Rubric-based evaluations | | | |
|---|---|---|---|---|---|---|
| 1 | Explain in detail the difference between both the terms data and information and support your answers with examples of each type? | 5 | (1 degree) The student's ability to introduce data, their role and shapes | (1 degree) The student's ability to introduce information, its upbringing and its use. | (2 degrees) The student's ability to conclude the difference between data and information | (1 degree) The student's ability to enhance his explanation of data and information with realistic, related examples |
| 2 | Explain at length the role of increasing micro-disciplines and increasing topics in influencing the information explosion revolution? | 5 | (2 degrees) The student's ability to explain the reasons for the increasing specializations and the subject. | (2 degrees) The student's ability to the role and influence of increasing specialization in the information revolution. | (1 degree) The student's ability to link the reasons for the emergence of modern science with the explosion of information | |
| 3 | Through what you learned in the course, mention the comprehensive definition of the term information science? | 5 | (2.5) The student's ability to define the faces and role of the Information Science Department. | (2.5) The student's ability to introduce the tasks of information science specialists since the establishment of information to the delivery to the beneficiary. | | |
| 4 | Explain the distinctions in roles and responsibilities among administrative levels in detail and provide illustrative examples. | 10 | (3 degrees) The student's ability to identify different management levels | (4 degrees) The student's ability to explain the difference between the tasks and duties of each administrative level | (3 degrees) The student's ability to learn about the hierarchical sequence of the tasks and roles of different management levels | |
| 5 | Mention three of the main benefits of cloud computing from a business perspective with an explanation? | 10 | (4 degrees) The student's ability to mention the three benefits | (3 degrees) The student's ability to explain each benefit extensively | (3 degrees) The student's ability to explain the benefits of cloud computing in business administration | |
| 6 | Through what you learned in the course, mention the comprehensive definition of the term information science? | 5 | (3 degrees) The student's ability to provide a comprehensive definition of the term information technology | (2 degrees) The student's ability to mention examples of information technology operations. | | |
| 7 | Talk about the layers of the atmosphere, mentioning the height and temperature of each layer. | 5 | (1 degree) The student's ability to mention the names of the five layers correctly | (2 degrees) The student's ability to explain each layer extensively | (1 degree) The student's ability to conclude the difference between the role of each layer (temperature and height) | (1 degree) The student's ability to arrange the layers according to their proximity to the ground |
| 8 | What do you think about the importance of the ozone layer? | 5 | (2 degrees) The student's ability to mention the role of the ozone layer in protecting the land | (2 degrees) The student's ability to explain the classes that have a role in protecting the earth. | (1 degree) The student's ability to know the basic role of the ozone layer | |
| 9 | What is the scientific definition of environmental chemistry? | 5 | (2 degrees) The student's ability to perform the term scientifically | (2 degrees) The student's ability to determine the aspects of environmental chemistry | (1 degree) The student's ability to mention the importance of environmental chemistry for human and life | |
| 10 | Define the term biotechnology? | 5 | (2.5 degrees) The student's ability to provide a comprehensive definition of the term biotechnology | (2.5 degrees) The student's ability to mention examples of biotechnology. | | |
| 11 | Discuss eating genetically modified fruits healthy or not? | 5 | (2 degrees) The student's ability to explain the components of the genetically modified fruits. | (2 degrees) The student's ability to explain the benefits and negatives of genetically modified fruits. | (1 degree) The student's ability to list the reasons that make genetically modified fruits acceptable. | |
| 12 | Five applications of biotechnology in the medical field with explanation? | 5 | (3 degrees) The student's ability to mention five vital technology applications in the field of medicine. | (2 degrees) The student's ability to mention a simple explanation of each type. | | |

Table A.3: Scoring Criteria for Determining the Final Score, Set by Course Directors

# Appendix B. Figures

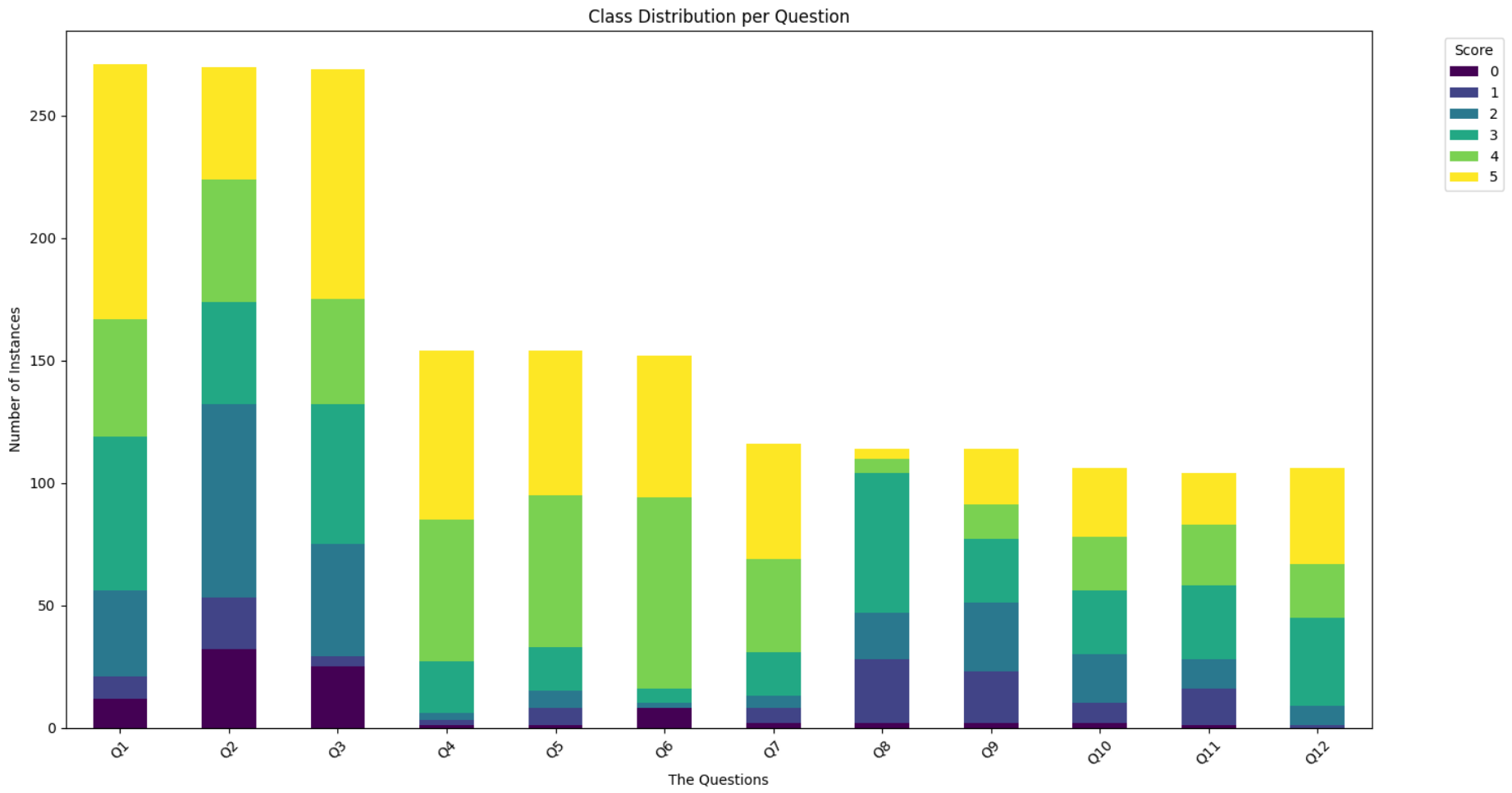

Figure B.1: Showing Class Distribution Across the Twelve Questions, with Scores Ranging from 0 to 5